\documentclass[12pt]{article}

\usepackage{latexsym,amsmath,amssymb}

\def\baq#1\eaq{\begin{align}#1\end{align}}
\def\baqn#1\eaqn{\begin{align*}#1\end{align*}}
\def\beq#1\eeq{\begin{equation}#1\end{equation}}
\def\beqn#1\eeqn{\begin{displaymath}#1\end{displaymath}}
\def\bqa#1\eqa{\begin{eqnarray}#1\end{eqnarray}}
\def\bqan#1\eqan{\begin{eqnarray*}#1\end{eqnarray*}}

\def\NNN{\mathbb N}
\def\RRR{\mathbb R}

\def\Expect{{\mathbf E}}
\def\Prob{{\mathbf P}}

\def\leqt{_{1:t}}
\def\ltt{_{<t}}

\def\leqT{_{1:T}}

\def\_norm{_\mathrm{norm}}

\newcommand{\zwidths}[1]{\rlap{$\scriptstyle #1$}}

\def\for_all{\mbox{ for all }}
\def\such_that{\mbox{ such that }}

\def\und{\mbox{ and }}

\newfont{\bbten}{bbold12}
\def\eins{{\hbox{\bbten 1}}}

\def\tfrac#1#2{{\textstyle\frac{#1}{#2}}}

\def\wwwtilde{$^{_{_\sim}}\!$}

\def\defaultskip{\medskip}

\newtheorem{Def1}{Definition}

\newtheorem{Prop}[Def1]{Proposition}
\newtheorem{Theorem}[Def1]{Theorem}

\newtheorem{Claim}[Def1]{Claim}

\newtheorem{Ex1}[Def1]{Example}

\newenvironment{Proof}[1][.]{\defaultskip\noindent \textbf{Proof}#1 }
  {\hspace*{0mm}\hfill $\Box$ \defaultskip}

\makeatletter
\newif\ifpdfoutput\@ifundefined{pdfoutput}%
  {\let\pdfoutput\@undefined}%
  {\ifcase\pdfoutput
     \let\pdfoutput\@undefined
   \else
     \pdfoutputtrue}
\makeatother
\ifpdfoutput
\fi

\newif\ifplatex\makeatletter\ifx\platex\@undefined
\platexfalse\else\platextrue\fi\makeatother
\ifplatex
\input{jdummy.def}
\fi

\usepackage{graphicx}

\newfont{\scL}{cmcsc10 scaled \magstep3}
\makeatletter
\ifcase\@ptsize 
\newfont{\scs}{cmcsc08}
\newfont{\its}{cmti08}
\or
\newfont{\scs}{cmcsc09}
\newfont{\its}{cmti09}
\or
\newfont{\scs}{cmcsc10}
\newfont{\its}{cmti10}
\fi
\@twosidetrue
\def\ps@myheadings{\let\@mkboth\@gobbletwo
\def\@oddhead{\hfil \hbox{}{\footnotesize{\rm\@Title}} \hfil
{\small{\rm\thepage}}}
\def\@oddfoot{}
\def\@evenhead{{\footnotesize{\rm
\thepage}}\hfil{\footnotesize{\sc\@Authors}}\hbox{} \hfil}
\def\@evenfoot{}
\def\sectionmark##1{}\def\subsectionmark##1{}}

\def\today{\ifcase\month\or
 January\or February\or March\or April\or May\or June\or
 July\or August\or September\or October\or November\or December\fi
 \space\number\day, \number\year}

\def\Authors#1{\gdef\@Authors{#1}}
\def\Title#1{\gdef\@Title{#1}}
\def\Number#1{\gdef\@Number{#1}}
\def\Phone#1{\gdef\@Phone{#1}}
\def\Fax#1{\gdef\@Fax{#1}}
\def\Name#1{\gdef\@Name{#1}}
\def\Date#1{\gdef\@Date{#1}}
\def\Series#1{\gdef\@Series{#1}}

\def\makecover{%
\begingroup
\@makecover 
\let\makecover\relax
 \let\@makecover\relax
\endgroup
}
\def\endmakecover{\if@restonecol\twocolumn\columnsep=3ex 
\else \newpage \fi}

\def\Titlebox{\vbox to 8.5cm}
\def\Coverbox{\vbox to 23.8cm}
\def\@makecover{
\@restonecolfalse\if@twocolumn\@restonecoltrue\onecolumn
 \else \newpage \fi 
\hsize=16.0truecm
\vsize=25.5truecm
\thispagestyle{empty}\c@page\z@
\protect\setlength{\parskip}{1ex plus .5ex minus
.5ex}\protect\setlength{\parindent}{1.3em}
\protect\setlength{\topmargin}{3cm}\protect\setlength{\hoffset}{-1.3cm}
\protect\setlength{\voffset}{-3.5cm}
\ifcase\@ptsize 
\setlength{\lineskip 1truept}\setlength{\normallineskip 1truept}
\def\baselinestretch{1}
\def\@normalsize{\@setsize\normalsize{20.3pt}\xivpt\@xivpt
\abovedisplayskip 12pt plus3pt minus7pt\belowdisplayskip \abovedisplayskip
\abovedisplayshortskip \z@ plus3pt\belowdisplayshortskip 6.5pt plus3.5pt
minus3pt\let\@listi\@listI}
\def\@listi{\leftmargin\leftmargini \parsep 4.5pt plus 2pt minus 1pt \itemsep
\parsep
 \topsep 9pt plus 3pt minus 5pt}
\def\large{\@setsize\large{25.2pt}\xivpt\@xivpt}
\def\Large{\@setsize\Large{30.8pt}\xviipt\@xviipt}
\def\LARGE{\@setsize\LARGE{35pt}\xxpt\@xxpt}
\def\huge{\@setsize\huge{42pt}\xxvpt\@xxvpt}
\def\small{\@setsize\small{19pt}\xipt\@xipt
\abovedisplayskip 11pt plus3pt minus6pt\belowdisplayskip \abovedisplayskip
\abovedisplayshortskip \z@ plus3pt\belowdisplayshortskip 6.5pt plus3.5pt
minus3pt
\def\@listi{\leftmargin\leftmargini \parsep 4.5pt plus 2pt minus 1pt \itemsep
\parsep
 \topsep 9pt plus 3pt minus 5pt}}
\normalsize
\setlength{\hoffset}{-2.1cm}
\or 
\setlength{\lineskip 1truept}\setlength{\normallineskip 1truept}
\def\baselinestretch{1}
\def\@normalsize{\@setsize\normalsize{20.3pt}\xivpt\@xivpt
\abovedisplayskip 12pt plus3pt minus7pt\belowdisplayskip \abovedisplayskip
\abovedisplayshortskip \z@ plus3pt\belowdisplayshortskip 6.5pt plus3.5pt
minus3pt\let\@listi\@listI}
\def\@listi{\leftmargin\leftmargini \parsep 4.5pt plus 2pt minus 1pt \itemsep
\parsep
 \topsep 9pt plus 3pt minus 5pt}
\def\large{\@setsize\large{25.2pt}\xivpt\@xivpt}
\def\Large{\@setsize\Large{30.8pt}\xviipt\@xviipt}
\def\LARGE{\@setsize\LARGE{35pt}\xxpt\@xxpt}
\def\huge{\@setsize\huge{42pt}\xxvpt\@xxvpt}
\def\small{\@setsize\small{19pt}\xipt\@xipt
\abovedisplayskip 11pt plus3pt minus6pt\belowdisplayskip \abovedisplayskip
\abovedisplayshortskip \z@ plus3pt\belowdisplayshortskip 6.5pt plus3.5pt
minus3pt
\def\@listi{\leftmargin\leftmargini \parsep 4.5pt plus 2pt minus 1pt \itemsep
\parsep
 \topsep 9pt plus 3pt minus 5pt}}
\normalsize
\setlength{\hoffset}{-1.8cm}

\or \relax \fi

\if@restonecol\onecolumn\vsize=26truecm\hsize=16.0truecm
\protect\setlength{\hoffset}{1ex}\else\relax\fi
\newfont{\siimfont}{msbm10 scaled \magstep5}
\newfont{\ssiimfont}{msbm10 scaled \magstep2}
\newcounter{siitrcounter}
\setcounter{siitrcounter}{\number\year}
\addtocounter{siitrcounter}{-2000}
\def\nn{\ifnum \number\year<2010 {\large\bf 0\arabic{siitrcounter}}\else
{\large\bf \arabic{siitrcounter}}\fi}
\def\mhhh{\mbox{-}}
\setcounter{page}{0}
\Coverbox{
\begin{flushright}
{\ssiimfont TCS}\hspace*{2pt}\mhhh{\ssiimfont TR}-{\ssiimfont
\@Series}-\nn-{\large\bf\@Number}
\end{flushright}
\begin{flushleft}
{\siimfont TCS}\hspace*{1ex} {\huge Technical Report}
\end{flushleft}
\par
\vskip 1in
\Titlebox{\vss
\begin{center}
{\LARGE \@Title \par}
\end{center}
\vss
\begin{center}
{\par {\large\bf by} \par}
\end{center}
\vskip 1.5ex
\begin{center}
{\global\baselineskip25truept\scL \@Authors}
\end{center}
\vss}
\vskip 3ex
\centerline{\large\bf Division of Computer Science}
\centerline{\large\bf Report Series \@Series}
\centerline{\large \@Date}
\par
\vskip 3ex
\begin{center}
\includegraphics*[width=2.5cm]{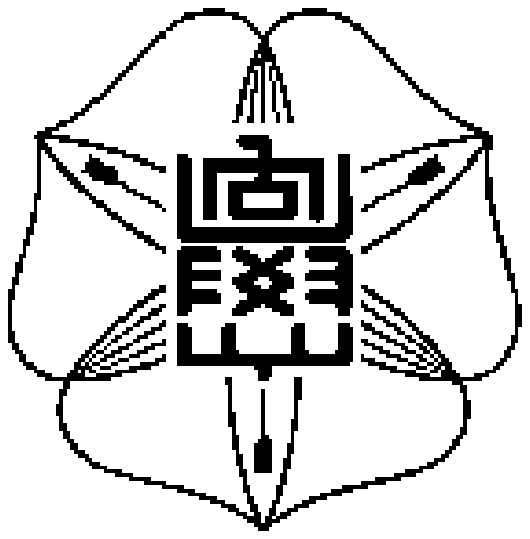}
\end{center}
\par
\vskip -.2cm
\par
\global\baselineskip16truept
\centerline{\huge Hokkaido University}
\centerline{\large Graduate School of}
\centerline{\large Information Science and Technology}
\par
\vskip 2ex
{\par \small 
\begin{minipage}[t]{8.26cm}
\begin{tabular}
{lr}Email:& \@Name @ist.hokudai.ac.jp\\ 
          & 
\end{tabular}
\end{minipage}
\begin{minipage}[t]{6.26cm}
\begin{flushright}
\begin{tabular}{lr}
Phone: & +81-011-\@Phone\\
Fax: & +81-011-\@Fax
\end{tabular}
\end{flushright}
\end{minipage}
}
\ifcase\@ptsize
\global\baselineskip12truept
\or
\global\baselineskip14truept
\or
\global\baselineskip16truept
\fi
}
\newpage
\pagenumbering{roman}
\setcounter{page}{1}
\thispagestyle{empty}
\Titlebox{}
$\phantom{this is an empty page}$
\newpage
\pagenumbering{arabic}
\endmakecover
\pagestyle{myheadings}
\ifplatex
\markboth{\hss{\footnotesize \@Authors}\hss}{\hss{\footnotesize \@Title}\hss}
\else
\markboth{\hss{\scs \@Authors}\hss}{\hss{\its \@Title}\hss}
\fi
}
\if\makecover\global\textheight=25.5cm \relax 
\else
\ifcase\@ptsize
\global\protect\textwidth 12.2cm
\global\protect
\global\parskip 1ex plus .5ex minus
.5ex\global\parindent1.3em\global\topmargin3cm
\global\hoffset=-.4cm\global\voffset=-1cm
\newfont{\bi}{cmbxti10}
\newfont{\bs}{cmbsy10}
\if@twocolumn \global\protect\textwidth 7.00in
\global\protect
\global\protect
\global\protect\columnsep=5ex
\protect\hoffset=-.9cm\global\voffset=-3.5cm
\renewenvironment{abstract}{\centerline{\bf
Abstract}\vspace{0.5ex}\begin{quote}\small}{\par\end{quote}\vskip 1ex}
\fi
\or
\global\protect\textwidth 13.75cm
\global\protect
\global\parskip 1ex plus .5ex minus
.5ex\global\parindent1.3em\global\topmargin3cm
\global\hoffset=-.8cm\global\voffset=-2.5cm
\newfont{\bi}{cmbxti10 scaled 1100}
\newfont{\bs}{cmbsy10 scaled 1100}
\if@twocolumn\global\protect\textwidth 7.00in
\global\protect
\global\protect
\global\protect\columnsep=5ex
\renewenvironment{abstract}{\centerline{\bf
Abstract}\vspace{0.5ex}\begin{quote}\small}{\par\end{quote}\vskip 1ex}
\fi
\or
\global\protect\textwidth 15.3cm
\global\parskip 1ex plus .5ex minus
.5ex\global\parindent1.3em\global\topmargin3cm
\global\hoffset=-1.1cm\global\voffset=-3.5cm
\newfont{\bi}{cmbxti10 scaled 1200}
\newfont{\bs}{cmbsy10 scaled 1200}
\if@twocolumn\global\protect\textwidth 7in
\global\protect
\global\protect
\global\protect\columnsep=5ex
\protect\hoffset=-.9cm\global\voffset=-3.5cm
\renewenvironment{abstract}{\centerline{\bf
Abstract}\vspace{0.5ex}\begin{quote}\small}{\par\end{quote}\vskip 1ex}
\fi
\fi
\fi
\Date{\today}
\makeatother

\Number{7}
\Authors{J. Poland}
\Title{FPL Analysis for Adaptive Bandits}
\Name{jan}
\Phone{706-7675}
\Fax{706-7675}
\Series{A}

\def\FPL{\mbox{FPL}}
\def\IFPL{\mbox{IFPL}}
\def\FPLt{\widetilde{\FPL}}
\def\IFPLt{\widetilde{\IFPL}}
\def\bFPL{\mbox{bFPL}}

\def\fplsmall{\mbox{\scriptsize\textup{F\kern-0.15emP\kern-0.15emL}}}
\def\fpl{^{\fplsmall}}
\def\fplt{^{\widetilde{\fplsmall}}}
\def\bfplsmall{\mbox{\scriptsize\textup{b\kern-0.12emF\kern-0.15emP\kern-0.15emL
}}}
\def\obfplsmall{\mbox{\scriptsize\textup{o\kern-0.1emb\kern-0.12em%
F\kern-0.15emP\kern-0.15em L }}}
\def\bfpl{^{\bfplsmall}}

\def\ifplt{^{\widetilde{\mbox{\scriptsize\textup{I\kern-0.08emF\kern-0.15emP\kern-0.15emL}}}}}
\def\scp{{\scriptscriptstyle^{\,\circ}}}

\begin{document}

\title{FPL Analysis for Adaptive Bandits}


\author{Jan Poland\thanks{
This work was supported by JSPS 21st century COE program C01.} \\
Grad.\ School of Inf.\ Sci.\ and Tech.\\
Hokkaido University, Japan \\
 \texttt{jan@ist.hokudai.ac.jp}\\
 \texttt{www-alg.ist.hokudai.ac.jp/\wwwtilde jan}}

\makecover

\maketitle

\begin{abstract}
A main problem of ``Follow the Perturbed Leader" strategies for online decision
problems is that regret bounds are typically proven against oblivious adversary.
In partial observation cases, it was not clear how to obtain
performance guarantees against adaptive adversary, without worsening the bounds.
We propose a conceptually simple argument to resolve this problem.
Using this, a regret bound of 
$O(t^{\frac{2}{3}})$ for FPL in the adversarial multi-armed bandit
problem is shown. This bound holds for the common FPL variant using only the
observations from designated exploration rounds. Using all observations allows
for the stronger bound of $O(\sqrt{t})$, matching the best
bound known so far (and essentially the known lower bound) for
adversarial bandits. Surprisingly, this variant does not even need explicit
exploration, it is self-stabilizing.
However the sampling probabilities have to be either
externally provided or approximated to
sufficient accuracy, using $O(t^2\log t)$ samples in each step.
\end{abstract}


\section{Introduction}\label{secIntro}

``Expert Advice" stands for an active research area which studies online algorithms. In each time step $t=1,2,3,\ldots$ the master algorithm, henceforth called \emph{master} for brevity, is required to commit to a decision, which results in some cost. The master has access to a class of \emph{experts}, each of which suggests a decision at each time step. The goal is to design master algorithms such that the \emph{cumulative regret} (which is just the cumulative excess cost) with respect to any expert is guaranteed to be small. Bounds on the regret are typically proven \emph{in the worst case}, i.e.\ without any statistical assumption on the process assigning the experts' costs. In particular, this might be an \emph{adaptive adversary} which aims at maximizing the master's regret and also knows the master's internal algorithm. This implies that (unless the decision space is continuous and the cost function is convex) the master must \emph{randomize} in order to protect against this danger.

In the recent past, a growing number of different but related online problems
have been considered. Prediction of a binary sequence with expert advice has
been popular since the work of Littlestone and Warmuth in the early 1990's.
Freund and Schapire \cite{Freund:97} removed the structural assumption on the
decision space and gave a very general algorithm called Hedge which in each time
step randomly picks one expert and follows its recommendation. We will refer to
this setup as the \emph{online decision problem}. Auer et al.\
\cite{Auer:95,Auer:03} considered the
first \emph{partial observation} case, namely the bandit setup, where in each time step the master algorithm only
learns its own cost, i.e.\ the cost of the selected expert. All these and many
other papers are based on \emph{weighted forecasting} algorithms.

A different approach, \emph{Follow the Perturbed Leader} (FPL), was pioneered as
early as 1957 by Hannan \cite{Hannan:57} and rediscovered recently by Kalai and
Vempala \cite{Kalai:03}. Compared to weighted forecasters, FPL has two main
advantages and one major drawback. First, it applies to the online decision
problem and admits a much more elegant analysis for \emph{adaptive learning
rate} \cite{Hutter:05expertx}. Even infinite expert classes do not cause much
complication. (However, the leading constant of the regret bound is
generically a factor of $\sqrt 2$ worse than that for weighted forcasters.)
Adaptive learning rate is necessary unless the total number of time steps to be
played is known in advance.

As a second advantage, FPL also admits efficient treatment of cases where the
expert class is potentially huge but has a linear structure
\cite{McMahan:04,Awerbuch:04}. We will refer to such problems as \emph{geometric
online optimization}. An example is the online shortest path problem on a graph,
where the set of admissible paths = experts is exponential in the number of
vertices, but the cost of each path is just the sum of the costs of the
vertices. 

FPL's main drawback is that its general analysis only applies against an
\emph{oblivious} adversary, that is an adversary that has to decide on
\emph{all} cost vectors before the game starts -- as opposed to an adaptive one
that before each time step $t$ just needs to commit to the current cost vector.
For the full information game, one can show that a regret bound against
oblivious adversary implies the \emph{same} bound against an adaptive one
\cite{Hutter:05expertx}. 
The intuition is that FPL's current decision at time $t$ does not depend on its
past decisions. Therefore, the adversary may well decide on the current cost
vector before knowing FPL's previous decisions. This argument does not apply in
partial observation cases, as there FPL's behavior does depend on its past
decisions (because the observations do so). As a consequence, authors started to
explicitly distinguish between oblivious and adaptive adversary, sometimes
restricting to the former, sometimes obtaining bounds of lower quality for the
latter. E.g.\ McMahan and Blum \cite{McMahan:04} suggest a workaround, proving
sublinear regret bounds against an adaptive bandit, however of worse order
($t^{\frac{3}{4}}\sqrt{\log t}$ instead of $t^{\frac{2}{3}}$, for both,
geometric online optimization and online decision problem). This is not
satisfactory, since in case of the bandit online decision problem for a suitable
weighted forecaster, even a $O(\sqrt t)$ bound against adaptive adversary is
known \cite{Auer:03}.

In this work, we remove FPL's major drawback. We give a simple argument 
(Section \ref{secFPL}) which shows that also in case of partial observation, a
bound for FPL against an oblivious adversary
implies the same bound for adaptive adversary. This will allow in particular to prove a
$O\big((tn\sqrt{\log n})^{\frac{2}{3}}\big)$ bound for the bandit online
decision problem (Section \ref{secOnline}). This bound is shown for the common
construction where only the observations of designated exploration rounds are
used. As this master algorithm is label efficient, the bound is essentially
sharp. In contrast, using all informations will enable us to prove a stronger
$O(\sqrt{tn\log n})$ bound (Section \ref{secUseAll}). This matches the best
bound known so far for the adversarial bandit problem \cite{Auer:03}, which
is sharp within $\sqrt{\log n}$. The downside of this algorithm is that
either the sampling probabilities have to be given by an oracle, or they have to
be approximated with to sufficient accuracy, using $O(t^2\log t)$ samples. 
The case of an infinite expert class is briefly discussed in Section
\ref{secInfinite}.

\section{FPL: oblivious $\Rightarrow$ adaptive}\label{secFPL}

Assume that $c_1,c_2,\ldots\in [0,1]^n$ is a sequence of cost vectors. There are
$n\geq 1$ experts. 
(We will give an example with infinitely many experts Section \ref{secInfinite},
but for simplicity of presentation, we restrict our main exposition to finite
expert classes). 
That is, $c_t^i$ is expert $i$'s cost at time $t$, and the costs are bounded
(w.l.o.g.\ in $[0,1]$). In the full observation game, at time $t$ the master
would know the past
cumulative costs $c\ltt=c_{1:t-1}=\sum_{s=1}^{t-1}c_s$ (observe that we have
introduced some notation here). 
However, our focus are \emph{partial observations} where this is not the case.
Hence, assume that there are \emph{estimates} $\hat c_t$ (to be specified later)
for the cost vectors $c_t$. Then at time $t$, $\FPL(t)$ samples a perturbation
vector $q_t\in[0,\infty)^n$ the components of which are independently
exponentially distributed, that is, $\Prob(q_t^i\geq x)=e^{-x}$. Afterwards,
the expert with the best (minimum) score $\hat c\ltt-\frac{q_t}{\eta_t}$ is
selected, where $\eta_t>0$ is the \emph{learning rate}:
\beq \label{eqFPL}
\FPL(t,\hat c\ltt)=\arg\min_{1\leq i\leq n}\Big\{\hat c^i\ltt-\tfrac{q^i_t}{\eta_t}\Big\}
\mbox{ where } q^i_t\stackrel{d.}{\sim}\mbox{Exp independently}.
\eeq
Denote the expert FPL chooses at time $t$ by $I_t=\FPL(t,\hat c\ltt)$.
Then an adaptive adversary is a function $A:[0,1]^{n\times t-1}\times \{1\ldots n\}^{t-1}\to[0,1]^n$. (We assume $A$ to be deterministic but remark that all our results and proofs hold for randomized $A$ without major modification.) The complete game between FPL and $A$ is specified by $c_t=A(c_1c_2\ldots c_{t-1},I_1I_2\ldots I_{t-1})$ and $I_t=\FPL(t,\hat c\ltt)$ for $t=1,2,\ldots$
The estimated cost vector $\hat c_t$ is revealed to FPL after time $t$ and specified by a mechanism ``outside" this game which is defined later (this is the exploration). 

After the game has proceeded for a number of time steps $T$, we want to evaluate FPL's performance. Actually, the \emph{expected} performance is the right quantity to address. If we are rather interested in high probability bounds on the actual performance, then they are easily obtained by observing that the difference of actual to expected performance is a martingale with bounded differences (all instantaneous costs $c_t^i$ are in $[0,1]$). Thus, high probability bounds follow by Azuma's inequality, as we will
demonstrate in Proposition \ref{propHP}.

How can we compute FPL's expected costs $\Expect c\leqT\fpl=\Expect\sum_{t=1}^T
c_t^{I_t}$? The key observation is that -- on the cost vectors generated by FPL
and $A$ and with the given estimated costs $\hat c_t$ -- FPL's expected costs
at time $t$ are the same as another algorithm $\FPLt$'s expected costs. $\FPLt$
is defined by
\beq \label{eqFPLt}
\FPLt(t,\hat c\ltt)=\arg\min_{1\leq i\leq n}\Big\{\hat c^i\ltt-\tfrac{q^i_*}{\eta_t}\Big\},
\eeq
where $q_*$ is a \emph{single fixed} vector with independently exponentially distributed components. Since we have to be careful to take expectations w.r.t.\ the appropriate randomness, we explicitely refer to the randomness in the notation by writing e.g.\ $\Expect c_t\fpl=\Expect_{q_t} c_t\fpl$. Then the following statement trivially holds, as $q_t$ and $q_*$ have the same distribution.

\begin{Prop} \label{propArgument}
At each time $t\leq T$, we have $\Expect_{q_t} c_t\fpl=\Expect_{q_*} c_t\fplt$.
\end{Prop}

This means that in order to analyze FPL, we may now proceed by considering the expected costs of $\FPLt$ instead. We can use the standard analysis based on the tools by Kalai and Vempala \cite{Kalai:03}, which requires that $\FPLt$ is executed on a sequence of cost vectors that is fixed and not known in advance. Actually, in contrast to the full observation game analysis, the bandit analysis will \emph{never} require the true cost vectors to be revealed, but rather the estimated cost vectors. For the cost vectors generated by $A$ in response to FPL, the prerequisite for $\FPLt$ is satisfied -- just consider $\FPLt$ as a \emph{virtual} or \emph{hypothetic} algorithm which is not actually executed. Therefore it does not make any decisions or cause any response from the adversary. Just for the sake of analysis we \emph{pretend} that it runs and evaluate the expected cost it incurs, which is the same as FPL.

Since our key argument and the way it is used in the analysis appears quite subtle at the first glance, we encourage the reader to thoroughly verify each of the subsequent formal steps. 

\section{The standard strategy against adversarial bandits}\label{secOnline}

\begin{figure}[t!]
\def\ii{\hspace*{1.5em}}
\begin{center}
\fbox{
\begin{minipage}{0.71\textwidth}
For $t=1,2,3,\ldots$\\
\ii set $\hat c_t^i=0$ for all $i$\\
\ii sample $r_t\in\{0,1\}$ independently s.t. $P[r_t=1]=\gamma_t$\\
\ii If $r_t=0$ Then set $I_t^b=\FPL(t,\hat c\ltt)$ according to (\ref{eqFPL})\\
\ii If $r_t=1$ Then sample $I_t^b$ from $\{1\ldots n\}$ uniformly ($I_t^b=u_t$)\\
\ii play decision $I_t^b$ and observe cost $c_t^{I_t^b}$\\
\ii If $r_t=1$ Then set $\hat c_t^{I_t^b}=n\cdot c_t^{I_t^b}/\gamma_t$
\end{minipage}}
\caption{The algorithm bFPL. The exploration rate $\gamma_t$ and the
learning rate $\eta_t$ (used by subroutine FPL) will be specified in Theorem
\ref{thMain}.}
\label{figbFPL}
\end{center}
\end{figure}

The first algorithm we consider, \emph{bandit-FPL} (bFPL), is specified in
Figure \ref{figbFPL} and proceeds as follows. At time $t$, it decides if to
perform an exploration or an exploitation step according to some
\emph{exploration probability} $\gamma_t\in(0,1)$. This is realized by sampling
$r_t\in\{0,1\}$ independently from all other randomness with
$P[r_t=1]=\gamma_t$. In case of exploration ($r_t=1$), the decision $I_t^b$ is
uniformly sampled from $\{1\ldots n\}$, independently from all other randomness.
We denote this choice by $u_t$. (For notational convenience, we will also refer
to the irrelevant $u_t$'s in the exploitations steps later.) In case of
exploitation ($r_t=0$), bFPL obtains its decision $I_t^b$ by invoking FPL
according to (\ref{eqFPL}). After bFPL has played its decision, it observes its
own costs $c_t^{I_t^b}$. Finally, only in case of exploration ($r_t=1$), the
estimated cost vector is set to something different from 0. This is the
\emph{standard way} of constructing an FPL variant against an adversarial
bandit \cite{McMahan:04,Awerbuch:04}. We will discuss how to make use of
\emph{all} observations in the
next section. Here is the formal specification of the algorithm again.
\beqn
I_t^b=\bFPL(t,\hat c\ltt)=\left\{\begin{array}{l l}
u_t & \mbox{if } r_t=1\\
\FPL(t,\hat c\ltt) & \mbox{otherwise,}\end{array}\right.\quad
\hat c_t^i=\left\{\begin{array}{l l}
\frac{nc_t^i}{\gamma_t} & \mbox{if } r_t=1\wedge i=I_t^b\\
0 & \mbox{otherwise.}
\end{array}\right. 
\eeqn
Consequently, the estimated cost vector is chosen \emph{unbiasedly}, i.e.\
$\Expect_{r_t,u_t} \hat c_t^i=c_t^i$. This technique was introduced in
\cite{Auer:95}.

\begin{Theorem} \label{thMain}
Let $\gamma_t=\min\big\{1,t^{-\frac{1}{3}}\big(n\sqrt{\log n}\big)^{\frac{2}{3}}\big\}$ and $\eta_t=\frac{\gamma_t}{n^2} t^{-\frac{1}{3}}\big(n\sqrt{\log n}\big)^{\frac{2}{3}}$. Then, for any $T\geq(n\log n)^2$, each expert $i\in\{1\ldots n\}$, and \emph{any} adaptive assignment of the costs $c_1,c_2,\ldots$, bFPL satisfies the regret bound
\beq\label{eqTh1toshow}
\Expect c\leqT\bfpl-c\leqT^i \leq 4\Big(Tn\sqrt{\log n}\Big)^{\frac{2}{3}}.
\eeq
(For $T<(n\log n)^2$, the regret is clearly at most $(n\log n)^2$.)
\end{Theorem}

\begin{Proof}
All computations we use in the subsequent proof have been taken or adapted from other work. Our point is to bring them into the right order and to carefully check that in this context, against an adaptive adversary, all operations are legitimate. In particular we have to take care that all expectations are w.r.t.\ the appropriate randomness. Again, we make this explicit in the notation and write e.g.\ $\Expect c_t\bfpl=\Expect_{q_t,r\leqt,u\leqt} c_t\bfpl$. Note that according to the definition of bFPL, $\Expect c_t\bfpl$ in fact does \emph{not} depend on $q\ltt$. During the proof, we will avoid the use of unspecified expectation (without subscripts). Let's introduce abbreviation $h\ltt=(r\ltt,u\ltt,q\ltt)$ for the randomization history, i.e. the tuple containing all past random variables.

Moreover, we will use \emph{conditional expectation}. For instance, $\Expect_{q_t}[c_t\fpl|h\ltt]$ denotes a random variable depending on the randomization history $h\ltt$, where for each possible history the expectation is taken w.r.t.\ $q_t$.  
Since we admit adaptive assignments, we must be aware that they may depend on
bFPL's past randomness. To make this explicit, we use the notation 
$\Expect [c_t^i|h\ltt]$ for the adversary's decisions and rewrite our bound to
show (\ref{eqTh1toshow}) as
\beq\label{eqTh1toshow1}
\sum_{t=1}^T \Expect_{q_t,r_t,u_t} [c_t\bfpl|h\ltt]-
\sum_{t=1}^T \Expect [c_t^i|h\ltt] \leq 4\Big(Tn\sqrt{\log n}\Big)^{\frac{2}{3}}.
\eeq
In order to keep the presentation simple, we assume the adversary to be
deterministic. Then for given randomization
history, $c_t^i$ is constant. The same proof (and hence the theorem) remains
valid if we admit randomized adversaries.

First note that $\Expect_{q_t,r_t,u_t} [c_t\bfpl|h\ltt]\leq \Expect_{q_t} [c_t\fpl|h\ltt]+\gamma_t$ holds in each time step $t$ by definition of bFPL and $c_t^{I_t^b}\leq 1$. Since
$\gamma_t\leq t^{-\frac{1}{3}}\big(n\sqrt{\log n}\big)^{\frac{2}{3}}$, we have
\beq\label{eqSumgamma}
\sum_{t=1}^T \gamma_t\leq
\sum_{t=1}^T
t^{-\frac{1}{3}}\big(n\sqrt{\log n}\big)^{\frac{2}{3}}
\leq \tfrac{3}{2}\Big(Tn\sqrt{\log n}\Big)^{\frac{2}{3}}.
\eeq
Therefore, (\ref{eqTh1toshow1}) follows from
\beq\label{eqTh1toshow2}
\sum_{t=1}^T \Expect_{q_t} [c_t\fpl|h\ltt]-
\sum_{t=1}^T \Expect [c_t^i|h\ltt] \leq 
\tfrac{5}{2}\Big(Tn\sqrt{\log n}\Big)^{\frac{2}{3}}.
\eeq
Consider this form of FPL (i.e.\ FPL executed in each time step) as a
\emph{virtual} algorithm: It does not run in that way on the inputs. Rather, for
the sake of analysis, we pretend that it runs with the $\hat c_t$ obtained from
bFPL and try to evaluate its (virtual) performance.

We then use Proposition \ref{propArgument} to bring into the play another virtual algorithm, namely $\FPLt$. Since for given randomization history, the expected performance of FPL and $\FPLt$ coincide, (\ref{eqTh1toshow2}) is proven if we can show 
\beq\label{eqTh1toshow25}
\sum_{t=1}^T \Expect_{q_*} [c_t\fplt|h\ltt]-
\sum_{t=1}^T \Expect [c_t^i|h\ltt]\leq 
\tfrac{5}{2}\Big(Tn\sqrt{\log n}\Big)^{\frac{2}{3}}.
\eeq 
Next, we perform the transition from real to estimated costs. Since the estimate $\hat c$ was defined to be unbiased, we have $\Expect[c_t^i|h\ltt]=\Expect_{r_t,u_t}[\hat c_t^i|h\ltt]$.
By the same argument, since the choice of $\FPLt$ actually does not depend on $r_t$ and $u_t$,
$ \Expect_{q_*} [c_t\fplt|h\ltt]=\Expect_{q_*,r_t,u_t} [\hat c_t\fplt|h\ltt]$
holds.
Hence, (\ref{eqTh1toshow25}) follows from
\beq\label{eqTh1toshow3}
\sum_{t=1}^T\Expect_{q_*,r_t,u_t} [\hat c_t\ifplt|h\ltt]-\sum_{t=1}^T \Expect_{r_t,u_t}[\hat c_t^i|h\ltt] \leq 
\tfrac{5}{2}\Big(Tn\sqrt{\log n}\Big)^{\frac{2}{3}}.
\eeq
Note that, somewhat curiously, $\FPLt$ (like FPL) only incurs estimated costs in
case of exploration, i.e.\ where it actually did not decide the action.
We need yet another virtual algorithm, \emph{infeasible} $\FPLt$ or $\IFPLt$, defined as
\beq \label{eqIFPLt}
\IFPLt(t,\hat c\leqt)=\arg\min_{1\leq i\leq n}\Big\{\hat c^i\leqt-\tfrac{q^i_*}{\eta_t}\Big\},
\eeq
which uses the same perturbation $q_*$ as $\FPLt$. It is not feasible because at
time $t$ it makes use of the information $\hat c_t$, which is only available
afterwards. As it is a virtual algorithm, this does not cause any problems. 
By \cite[Theorem 4]{Hutter:05expertx}, which is proven by an argument very
similar to (\ref{eq:fplifpl}) below, in case of 
exploration (i.e.\ $r_t=1$) it holds that 
$\Expect_{q_*} [\hat c\fplt_t|h\ltt,r_t=1]\leq \Expect_{q_*} [\hat c\ifplt_t|h\ltt,r_t=1]+\eta_t\big(\frac{n}{\gamma_t}\big)^2$.
We remark that this step is valid also for independently sampled
perturbations $q_t$. Clearly, 
$\Expect_{q_*} [\hat c\fplt_t|h\ltt,r_t=0]=\Expect_{q_*} 
[\hat c\ifplt_t|h\ltt,r_t=0]$ 
in case of exploitation ($r_t=0$).
Thus in expectation w.r.t.\ $q_*$ and $r_t$, and for any $u_t$,
\beqn
\Expect_{q_*} [\hat c_t\fplt|h\leqT]=\Expect_{q_*,r_t} [\hat c\fplt_t|h\ltt]\leq \Expect_{q_*,r_t} [\hat c\ifplt_t|h\ltt]+\tfrac{\eta_t n^2}{\gamma_t}.
\eeqn
The sum over $\frac{\eta_t n^2}{\gamma_t}\leq t^{-\frac{1}{3}}\big(n\sqrt{\log n}\big)^{\frac{2}{3}}$ is bounded as in (\ref{eqSumgamma}), and we see that (\ref{eqTh1toshow3}) holds if we can show
\beq\label{eqTh1toshow4}
\sum_{t=1}^T\Expect_{q_*,r_t,u_t} [\hat c_t\ifplt|h\ltt]-
\sum_{t=1}^T \Expect_{r_t,u_t}[\hat c_t^i|h\ltt] \leq 
\Big(Tn\sqrt{\log n}\Big)^{\frac{2}{3}}.
\eeq
The rest of the proof now follows as in \cite{Kalai:03} or \cite{Hutter:05expertx}. In order to maintain self-containedness, we give it here. Actually we verify (\ref{eqTh1toshow4}) for \emph{any} choice of $r\leqT,u\leqT$, then it also holds in expectation. 

In the following, we suppress the dependency on $r\leqT,u\leqT$ in the notation.
Then all expectations are w.r.t.\ $q_*$.
We use the following convenient notation from \cite{Kalai:03}: For a vector $x\in\RRR^n$, let $M(x)$ be the unit vector which has a 1 at the index $\arg\min_i\{x^i\}$ and 0's at all other places. Then the process of selecting a minimum can be written as scalar product: $\min_i\{x^i\}=M(x)\scp x$. For convenience, let $\eta_0=\infty$ and $\tilde c\leqt=\hat c\leqt-\frac{q_*}{\eta_t}$. Then it is easy to prove by induction \cite{Kalai:03,Hutter:05expertx} that
\beq\label{eqnoregret}
  \hat c\ifplt\leqt-\sum_{t=1}^T M(\tilde c_{1:t})\scp q_*\Big(\tfrac{1}{\eta_t}-\tfrac{1}{\eta_{t-1}}\Big)=
  \sum_{t=1}^T M(\tilde c_{1:t})\scp \tilde c_t \leq M(\tilde c_{1:T})\scp \tilde c_{1:T}.
\eeq
In order to estimate $\Expect\hat c\ifplt\leqt$, we take expectations on both sides. Then observe $\Expect M(\tilde c_{1:T})\scp \tilde c_{1:T}\leq \Expect M(\hat c_{1:T})\scp \tilde c_{1:T}=\min_j\{\hat c_{1:T}^j\}-\frac{\Expect M(\hat c_{1:T})\scp q_*}{\eta_T}\leq \mbox{$\hat c_{1:T}^i-\frac{1}{\eta_T}$}$ by definition of $M$. The negative term on the l.h.s.\ of (\ref{eqnoregret}) may be bounded by $\sum_{t=1}^T M(\tilde c_{1:t})\scp q_*\Big(\tfrac{1}{\eta_t}-\tfrac{1}{\eta_{t-1}}\Big)\leq\sum_{t=1}^T M(-q_*)\scp q_*\Big(\tfrac{1}{\eta_t}- \tfrac{1}{\eta_{t-1}}\Big)=\frac{\max_i\{q_*^i\}}{\eta_T}\leq\frac{1+\log n}{\eta_T}$ (see \cite{Kalai:03} or \cite{Hutter:05expertx} for the last estimate). Plugging these estimates back into (\ref{eqnoregret}) while observing $\tfrac{1}{\eta_t}-\tfrac{1}{\eta_{t-1}}\geq 0$ and $\eta_T=T^{-\frac{2}{3}}\big(\frac{\log n}{n}\big)^{\frac{2}{3}}$ (which holds because of $T\geq(n\log n)^2$), finally shows (\ref{eqTh1toshow4}) and concludes the proof of the theorem.
\end{Proof}

\begin{Prop} \label{propHP} (High probability bound)
For each $T\geq 1$ and $0\leq\delta\leq 1$, the actual costs of bFPL are bounded
with probability at least $1-\delta$ by
\beqn
c\leqT\bfpl\leq\Expect c\leqT\bfpl+\sqrt{2T\log\tfrac{2}{\delta}}.
\eeqn
\end{Prop}
\begin{Proof}
Again we use the explicit notation from the proof of the previous theorem.
It is easy to see that the sequence of random variables
$X_T=\sum_{t=1}^T\big(c_t\bfpl-\Expect_{r_t,u_t,q_t}[c_t\bfpl|h\ltt]\big)$
is a martingale w.r.t.\ the filter of sigma-algebras generated by the
randomization history $h\leqt$. Moreover, its differences are bounded by
$|X_t-X_{t-1}|\leq 1$. Consequently, by Azuma's inequality, the probability that
$X_t$ exceeds some $\lambda>0$ is bounded by
$\delta=2\exp\big(-\frac{\lambda^2}{2T}\big)$. Solve this for $\lambda$ to
obtain the assertion.
\end{Proof}

\section{Using all observations}\label{secUseAll}

The algorithm bFPL considered so far does only uses a $\gamma$-fraction of all
the input. It is thus a \emph{label efficient} decision maker
\cite{Cesa:04,Cesa:04partial}. One possible way to specify a label efficient
problem setup is to require that the master usually does not observe
anything, and it incurs maximal cost if it decides to observe something
\cite{Cesa:04partial}. Since just before (\ref{eqSumgamma}), we upper
bounded the costs in case of exploration by 1, it is 
immediate that the same analysis and hence also
Theorem \ref{thMain} transfer to the label efficient case. 
\cite[Sec.~5]{Cesa:04partial} prove that there is a label
efficient prediction problem such that \emph{any} forecaster
incurs a regret proportional to $t^{\frac{2}{3}}$. Hence the bound in
Theorem \ref{thMain} is essentially sharp for bFPL.

Of course, the usual bandit setup does not require the master to make use of
only a tiny fraction of all information available. For weighted forecasters, it
is very easy to produce an unbiased cost estimate if each round's inputs are
used. It turns out that then regret bound proportional to $\sqrt t$ can be
obtained \cite{Auer:03}. Unfortunately this is different for FPL, as here the
sampling probabilities are not explicitely available. In the following, we will
first 
discuss the computationally infeasible case assuming 
that we know the sampling probabilities. 
After that, we show how to approximate
them by a Monte Carlo simulation to sufficient accuracy.

Surprisingly, it is possible to work with the plain FPL algorithm from
(\ref{eqFPL}), without exploration. We just have to use the correct
estimated cost vectors,
\beq
\label{eq:chati}
\hat c_t^i=\left\{\begin{array}{l l}
c_t^i/\Prob(I_t\fpl=i) & \mbox{if } i=I_t\fpl\\
0 & \mbox{otherwise,}
\end{array}\right. 
\eeq
where $I_t\fpl$ was FPL's choice at time $t$. We assume that the values
$\Prob(I_t\fpl=i)$ are provided by some oracle. 

It is not hard to adapt the proof of Theorem \ref{thMain} to analyze FPL under
these conditions.
As in the steps up to (\ref{eqTh1toshow3}),
\beqn
\Expect_{q_t}[c_t\fpl|h\ltt]= 
  \Expect_{q_*,q_t,r_t,u_t} [\hat c_t\fplt|h\ltt]=
  \Expect_{q_*,q_t,r_t,u_t} [\hat c_t^{\widetilde{\fplsmall}(q_*)}(q_t)|h\ltt]. 
\eeqn
The overly explicit notation $\hat c_t^{\widetilde{\fplsmall}(q_*)}(q_t)$
serves to remind that the cost vector estimated is obtained using $q_t$, while
$\FPLt$'s choice incurring cost stems from $q_*$. It is essential that $q_t$ and
$q_*$ are independent. Observe that in general, 
$\Expect_{q_*,q_t,r_t,u_t} [\hat c_t^{\widetilde{\fplsmall}(q_*)}(q_t)|h\ltt]
\lneqq
\Expect_{q_t,r_t,u_t} [\hat c_t^{\widetilde{\fplsmall}(q_t)}(q_t)|h\ltt]$: 
the latter quantity, which is the actual estimated cost of $\FPLt$'s choice, is
biased and too large.

Abbreviate $p^i=\Prob(I\fplt_t=i)$ and
$\pi^i=\Prob(I\ifplt_t=i)$. Denote the
exponential distribution by $\mu$ and integration with respect
to $q^1\ldots q^n$ without the $i$th coordinate by 
$\int\ldots d\mu(q^{\neq i})$. Moreover, for $x\in\RRR$, 
let $x^+=\max\{x,0\}$. Then, similarly to the proof of
\cite[Theorem 4]{Hutter:05expertx},
\bqa
\label{eq:fplifpl}
p_i & = & 
  \int\int\limits_{\max\limits_{j\neq
  i}\zwidths{\{\eta_t(\hat c\ltt^i-\hat c\ltt^j)+q^j\}}}^\infty
  d\mu(q^i) d\mu(q^{\neq i})=
\int e^{-(\max\limits_{j\neq
i}\{\eta_t(\hat c\ltt^i-\hat c\ltt^j)+q^j\})^+}\!
   d\mu(q^{\neq i})\\ \nonumber
& \leq & \int e^{\frac{\eta_t}{p^i}}e^{-(\max\limits_{j\neq
i}\{\eta_t(\hat c\ltt^i-\hat c\ltt^j)+q^j\}+\frac{\eta_t}{p^i})^+}
   d\mu(q^{\neq i})\\ \nonumber
& \leq & e^{\frac{\eta_t}{p^i}} \int e^{-(\max\limits_{j\neq
i}\{\eta_t(\hat c\leqt^i-\hat c\leqt^j)+q^j\})^+}
   d\mu(q^{\neq i})
= e^{\frac{\eta_t}{p^i}}\pi^i.
\eqa
Hence, $\pi^i\geq p^i e^{-\frac{\eta_t}{p^i}}\geq 
p^i\left(1-\frac{\eta_t}{p^i}\right)=p^i-\eta_t$, which implies
\bqan
\Expect_{q_*,q_t,r_t,u_t} [\hat c_t\fplt|h\ltt]&=&
\sum_{i=1}^n p^i \sum_{j=1}^n p^j \eins_{i=j} \frac{c_t^i}{p^i}
= \sum_{i=1}^n p^i c_t^i\\
& \leq &
\sum_{i=1}^n \pi^i c_t^i+
n\eta_t
=\Expect_{q_*,q_t,r_t,u_t} [\hat c_t\ifplt|h\ltt]+
n\eta_t.
\eqan
This shows the step from feasible to infeasible FPL. The last step from
infeasible FPL to the best decision in hindsight proceeds as shown already above
and in \cite{Kalai:03,Hutter:05expertx}. Like
before, it causes the upper bound of the cumulative regret to increase by
$\frac{\log n}{\eta_T}$. This is true for any $(q\leqT,r\leqT,u\leqT)$, hence
also in expectation. The total regret is thus upper bounded by 
$\frac{\log n}{\eta_T}+n\sum_{t=1}^T\eta_t$, and we have just proved:

\begin{Theorem} \label{thChati}
The algorithm FPL (\ref{eqFPL}), obtaining cost estimates according to 
(\ref{eq:chati}) and with learning rate $\eta_t=\sqrt{\frac{\log n}{2nt}}$
achieves a regret of at most
\beq\label{eq:boundchati}
\Expect c\leqT\fpl-c\leqT^i \leq 2\sqrt{2Tn\log n} \mbox{\quad for any }
i\in\{1\ldots n\}.
\eeq
\end{Theorem}

We would like to point to a quite remarkable symmetry break here. It is
straightforward to formulate FPL and the analysis from Section
\ref{secOnline} for \emph{reward maximization} instead of cost minimization.
Then the (perturbed) leader is the expert with the highest (perturbed) reward, and perturbations are added to the
scores. In the full information game, this reward maximization is perfectly
symmetric to cost minimization by just setting $\mathit{reward}_t^i=1-c_t^i$:
all probabilities, distributions, and outcomes will be exactly the same. This
is different in the partial observation case: There, in case of reward, the
expert by FPL is the only one which can gain score. This is
an advantage, in contrast to the disadvantage in case of
loss minimization: Here, the selected expert is the only one to worsen its
score. Put it differently, there is an automatic exploration or
self-stabilization in the cost minimization case. With this intuition, it is
less surprising that we did not need explicit exploration in Theorem
\ref{thChati}. The corresponding result for reward maximization would not hold,
as simple counterexamples show. Formally, it is the step from FPL to
infeasible FPL which fails: A computation similar to (\ref{eq:fplifpl}) only
shows $\pi^i\leq p^i e^{\frac{\eta_t}{p^i}}$, which does not imply a sufficiently
strong assertion in general. However, reintroducing the exploration rate $\gamma_t$,
we may set $\eta_t=\frac{\gamma_t}{n}$. This implies $\frac{\eta_t}{p^i}\leq 1$
for all $i$, hence $e^{\frac{\eta_t}{p^i}}\leq1+2\frac{\eta_t}{p^i}$.
Letting $\gamma_t=\sqrt{\frac{n\log n}{t}}$, we can conclude a bound like
(\ref{eq:boundchati}).

\subsection{A computationally feasible algorithm}

We conclude this section by discussing a computationally feasible
variant of FPL using all observations. This algorithm is constructed in a
straightforward way: Select the current action $i=I_t\fpl$ according to FPL and
substitute the estimate $\hat c_t^i$ from (\ref{eq:chati}) by
$\hat c_t^i=\frac{c_t^i}{\hat p_t^i}$. It remains to estimate 
$\hat p_t^i$ by a Monte Carlo simulation.

There are two possibilities of error: either $\hat p_t^i$ overestimates 
$p_t^i$, or it underestimates $p_t^i$. The respective consequences are
different: If $\hat p_t^i>p_t^i$, then the instantaneous cost of the
selected expert is just underestimated. We can account for this by adding a
small correction to the instantaneous regret. At the end of the game, we
perform well with respect to the underestimated costs, which are  upper
bounded by the true costs. This does not cause any further problems.
The case $\hat p_t^i<p_t^i$ is more critical, since then at the end of the game
we perform well only w.r.t.\ overestimated costs. We therefore have to treat
this case more carefully.

Problems arise if the true probability $p_t^i$ is very close to 0, as then the
Monte Carlo sample might contain very few or no hits and the variance of the
estimated cost is high. Since FPL does not prevent this
case, we reintroduce $\gamma_t$ as an ``exploration threshold". Let
$\gamma_t=\frac{1}{2\sqrt t}\leq\frac{1}{2}$. We first assume that
$p^i_t\geq\gamma_t$. If this assumption is false but we use 
$\hat p^i_t\geq\gamma_t$, then $\hat p_t^i$ is an overestimate and we have to
consider an additional instantaneous regret. This case has probability
at most $\gamma_t$. Consequently, as (true) instantaneous costs are always
bounded by 1, the additional instantaneous regret is at most
$\gamma_t$.

We sample the perturbed leader $k\in\NNN$ times and denote by $a^i(k)$
the number of times the leader happens to be expert $i$. Recall that expert
$i$ is the one already selected by FPL. By Hoeffding's
inequality, the distribution of $\frac{a^i(k)}{k}$ is sharply peaked around its
mean $p^i$:
\beqn
\Prob\Big[\tfrac{a^i(k)}{k}-p^i\geq\tfrac{\gamma_t^2}{\sqrt 2}\Big]
\leq e^{-\gamma_t^4k} \und
\Prob\Big[\tfrac{a^i(k)}{k}-p^i\leq-\tfrac{\gamma_t^2}{\sqrt 2}\Big]
\leq e^{-\gamma_t^4k}.
\eeqn
We choose $k$ such that the probability bounds on the r.h.s.\ are at most
$\gamma_t$, i.e.\ $e^{-\gamma_t^4k}\leq\gamma_t$.
Consequently we should sample 
$k=\big\lceil\gamma_t^{-4}\log(\gamma_t^{-1})\big\rceil
=\big\lceil2t^{2}\log(2\sqrt{t})\big\rceil$ times. Hence the sampling complexity
of the algorithm is $O(t^2\log t)$.
Let
\beqn
\hat p_t^i:=\max\big\{\gamma_t,\tfrac{a^i(k)}{k}-
\tfrac{\gamma_t^2}{\sqrt 2}\big\},
\eeqn
then $\hat p_t^i\leq p_t^i$ with probability at least $1-\gamma_t$ (recall the
assumption $p^i_t\geq\gamma_t$). Hence the possibility of overestimate $\hat
p_t^i>p_t^i$ causes an additional regret of $\gamma_t$.

Finally we need to deal with possible underestimates. For some integer 
$m\geq 1$, the probability that $\hat p_t^i$ falls below 
$p_t^i-\frac{(\sqrt m+1)\gamma^2}{\sqrt 2}$ is at most
\beq
\label{eqSIP}
\Prob\Big[\tfrac{a^i(k)}{k}-p^i\leq-\tfrac{\sqrt m\gamma_t^2}{\sqrt 2}\Big]
\leq e^{-m\gamma_t^4k}\leq\gamma_t^m
\eeq
by Hoeffding's inequality. We partition the interval $[\gamma_t,p_i^t)$ of all
possible underestimates into subintervals 
$A_1=\big[p_t^i-\frac{2\gamma^2}{\sqrt 2},p_t^i\big)$ and
\beqn
A_m=\left[p_t^i-\tfrac{(\sqrt m+1)\gamma^2}{\sqrt 2},
p_t^i-\tfrac{(\sqrt{m-1}+1)\gamma^2}{\sqrt 2}\right),\quad m\geq 2.
\eeqn 
We do not need to consider $m$ with the property
$A_m\cap[\gamma_t,p_i^t)=\emptyset$. That is, we can restrict to $m$ small
enough that
$p_i^t-\sqrt{\frac{1}{2}}(\sqrt m+1)\gamma_t^2\geq
\gamma_t-\sqrt{\frac{1}{2}}\gamma_t^2$. Let $M$ be the largest $m$ for which
this condition is satisfied, then one can easily see
$\sqrt m+1\leq\sqrt M+1\leq\sqrt 2 
(p-\gamma_t+\sqrt{\frac{1}{2}}\gamma_t^2)/\gamma_t^2$.
\begin{Claim}
If $m\leq M$, then
$\frac{c_t^i}{p_t^i-(\sqrt m+1)\gamma_t^2/\sqrt{2}}\leq
\frac{c_t^i}{p_t^i}+\gamma_t(\sqrt m+1)$.
\end{Claim}
This follows by a simple algebraic manipulation. 
Consequently, for $\hat p_i^t\in A_m$, 
we have $\Expect \hat c_t^i\leq c_t^i+(\sqrt m +1)\gamma_t$.
Moreover, $\hat p_t^i\in A_m$ occurs with probability at most
$\gamma_t^{m-1}$ according to (\ref{eqSIP}). By bounding the expectation over
all $A_m$, we thus obtain an additional regret of at most
\beqn
\sum_{m=1}^{M} (\sqrt m +1)\gamma_t^m\leq
\gamma_t\sum_{m=0}^\infty (m +2)\gamma_t^m\leq
\frac{2\gamma_t}{1-\gamma_t}+
\frac{\gamma_t^2}{(1-\gamma_t)^2}\leq 5\gamma_t,
\eeqn
since $\gamma_t\leq\frac{1}{2}$. Altogether, this proves the following theorem.

\begin{Theorem} 
Let $\gamma_t=\frac{1}{2\sqrt t}$ be the exploration threshold. In each time
step, after selecting one expert $i$, let FPL obtain an estimate 
$\hat p_t^i=\max\big\{\gamma_t,\tfrac{a^i(k)}{k}-
\tfrac{\gamma_t^2}{\sqrt 2}\big\}$
for $\Prob(I_t\fpl=i)$, by sampling the perturbed leader
$k=\big\lceil2t^{2}\log(2\sqrt{t})\big\rceil$
times and counting the number of hits $a^i(k)$. 
Let the estimated cost of the selected
expert be $\hat c_t^i=c_t^i/\hat p_t^i$, and the estimated cost of all other
experts be zero. Then the algorithm FPL (\ref{eqFPL}) with learning rate
$\eta_t=\sqrt{\frac{\log n}{2nt}}$
achieves a regret of at most
\beq
\Expect c\leqT\fpl-c\leqT^i \leq 2\sqrt{2Tn\log n} +7\sqrt T 
\mbox{\quad for any } i\in\{1\ldots n\}.
\eeq
\end{Theorem}

\section{Infinite expert classes}\label{secInfinite}

\begin{figure}[t!]
\def\ii{\hspace*{1.5em}}
\begin{center}
\fbox{
\begin{minipage}{0.95\textwidth}
For $t=1,2,3,\ldots$\\
\ii set $\hat c_t^i=0$ for $i\in\{i:t\geq\tau^i\}$ and $\hat c_t^i=\big(\gamma_t\min\{w^i:t\geq\tau^i\}\big)^{-1}\!\!\!\!\!$ for $i\notin\{i:t\geq\tau^i\}$\\
\ii sample $r_t\in\{0,1\}$ independently s.t. $P[r_t=1]=\gamma_t$\\
\ii If $r_t=0$, set $I_t^b=\arg\min_{i:t\geq\tau^i}\big\{\hat c^i\ltt+\frac{\log w^i-q^i_t}{\eta_t}\big\}$ (FPL on the active experts)\\
\ii If $r_t=1$, sample $I_t^b\in\{i:t\geq\tau^i\}$ according to the weights $\frac{w^i}{\sum_{i:t\geq\tau^i}w^i}$\\
\ii play decision $I_t^b$ and observe cost $c_t^{I_t^b}$\\
\ii If $r_t=1$, set $\hat c_t^{I_t^b}=\big[c_t^{I_t^b}\sum_{i:t\geq\tau^i}w^i\big]/
\big[\gamma_tw^{I_t^b}\big]$
\end{minipage}}
\caption{The algorithm bFPL for infinite expert class. The entering times $\tau^i$, the exploration rate $\gamma_t$, and the learning rate $\eta_t$, will be specified in Theorem \ref{thInf}.}
\label{figbFPLinf}
\end{center}
\end{figure}

Here, we sketch a variant of bFPL, taken from
\cite{Poland:05dule}, with guaranteed worst-case
performance against a bandit with countably infinitely many
arms. So we consider the following setup: The adversary
subsequently generates cost vectors $c_t\in[0,1]^\infty$,
and at each time $t$ we have to select one index or expert
$i$ and incur its cost $c_t^i$. We learn only the cost of
the selected expert. 

As a prerequisite, we need that each of the infinitely many
experts is associated with a \emph{prior weight} $w^i$ such
that $\sum_i\leq 1$. Since in order to obtain a cost
estimate $\hat c$, the observed cost is divided by the
weight of the sampled expert, we have to be careful not to
admit too small weights. We need to keep control of the
maximum possible expected cost, since otherwise the step
from FPL to IFPL would be problematic. 
One possibility to do so is defining an \emph{entering time}
$\tau^i\geq 1$ for each expert. Prior to $\tau^i$, the
expert is not active and cannot be chosen. We choose
$\tau^i=\left\lceil\big(\frac{1}{w^i}\big)^{\frac{1}{\alpha}
}\right\rceil$, with $0<\alpha<1$ to be defined later. Then
it is not hard to see that the minimum weight of any active
expert at time $t$ is lower bounded:
$\min\{w^i:t\geq\tau^i\}\geq t^{-\alpha}$. Letting the
exploration rate be $\gamma_t=t^{-\beta}$ with $0<\beta<1$
to be defined later, the maximum unbiasedly estimated cost
is at most $t^{\alpha+\beta}$. For the step from FPL to IFPL
to go through, we thus may choose
$\eta_t=t^{-2\alpha-2\beta}$. Then both steps from bFPL to
FPL and from FPL to IFPL each cause a regret of at most
$\sum_{t=1}^T t^{-\beta}\leq\frac{1}{1-\beta}T^{1-\beta}$.
On the other hand, $\frac{1}{\eta_T}$ causes a regret of at
most $T^{2\alpha+2\beta}$. In order to minimize these bounds
simultaneously, we choose $\beta=\frac{1-2\alpha}{3}$.

In order to guarantee that the step from IFPL to some fixed
expert holds, we have to correctly assign estimated costs to
inactive experts. For example, if an expert enters the game
and previously has been assigned no estimated cost at all,
then a bound w.r.t.\ this expert may be difficult to obtain.
We therefore assign \emph{maximum possible} estimated costs
to all inactive experts. Then one can show
\cite{Poland:05dule} that, evaluating the expected costs,
the step IFPL to some fixed reference expert holds almost
without modification. Clearly, the reference expert's
estimated costs now exceed its true costs by at most 
$\sum_{t=1}^{\tau^i-1}t^{\alpha+\beta}$, which is easily
shown to be upper bounded by
$\big(\frac{1}{w^i}\big)^{1+\frac{1}{\alpha}+\frac{\beta}
{\alpha}}$. This gives another \emph{additive} bound to the
regret in terms of the weight of the reference expert --
there is not multiplicative factor of $\frac{1}{w^i}$ any
more. This is an artifact of 
the design of the algorithm and proof technique and does not
mean that the new variant performs better than the old one.
Actually, since $\alpha>0$, the bound is now
$O\big(t^{\frac{2}{3}+\frac{2\alpha}{3}}\big)$ as opposed to
$O\big(t^{\frac{2}{3}}\big)$ before. Choosing a large
$\alpha$ results in a small $\big(\frac{1}{w^i}\big)$ term,
but the order in $t$ gets large, while a small $\alpha$ has
the opposite effect.  

The complete algorithm is specified in Figure
\ref{figbFPLinf}. The following statement, which improves on
the bounds given in \cite{Poland:05dule} (they are based
on the workaround from \cite{McMahan:04}) is an example
where we select $\alpha=\frac{1}{8}$. 

\begin{Theorem}\label{thInf}
Consider a bandit problem with countably many arms/experts,
each expert $i$ having a prior weight $w^i$ such that the
weights sum up to at most $1$.
Then the above described bFPL variant with entering times
$\tau^i=\left\lceil\big(\frac{1}{w^i}\big)^8\right\rceil$,
exploration rate $\gamma_t=t^{-\frac{1}{4}}$, and learning
rate $\eta_t=t^{-\frac{3}{4}}$, satisfies the regret bound
\beqn
\Expect c\leqT\bfpl-c\leqT^i \leq
O\left(\Big(\tfrac{1}{w^i}\Big)^{11}+T^{\frac{3}{4}}\log
w^i\right)
\eeqn
for all $T\geq 1$, any adaptive assignment of the cost
vectors and any reference expert $i$.
\end{Theorem}

The formal proof is omitted. It follows the outline of Theorem \ref{thMain},
using the arguments discussed above. Many of the arguments, including the step
from IFPL to the reference expert, are formally carried out in
\cite{Poland:05dule}.

\section{Discussion}\label{secDC}

The main statement of this paper is the following:

\vspace{1ex}
\noindent
\hspace*{0.1\textwidth}
\begin{minipage}{0.8\textwidth}
\itshape
If we have a regret
minimization algorithm with a bound guaranteed against an oblivious adversary,
and if the algorithm chooses the current action/expert by some
independent random sampling based on past cumulative scores (e.g.\ FPL or
weighted majority), then the same bound also holds against an adaptive
adversary. This is true both for full and partial observations.
\end{minipage}
\vspace{1ex}

\noindent
We have used this argument for showing bounds for FPL in the adversarial bandit
problem. The strategy to use only feedback from exploration rounds which is
common for FPL achieves a regret bound of $O(t^{\frac{2}{3}})$. As the
algorithm is label efficient, this bound is sharp. Using all observations
allows to push the regret down to $O(\sqrt t)$. Then however the sampling
probabilities have to be approximated.

In the same way, it is possible to use our argument for the general geometric
online optimization problem \cite{McMahan:04,Awerbuch:04}, also resulting in a
$O(t^{\frac{2}{3}})$ regret bound against adaptive adversary. An interesting
open problem is the following: Under which conditions and how is it possible to
use all observations in the geometric online optimization problem, hopefully
arriving at a $O(\sqrt t)$ bound?

We conclude with a note on \emph{regret against an adaptive adversary}. We
considered the \emph{external} regret w.r.t.\ the best action/strategy/expert
from a pool. There are two directions from here. One is to go to different
regret definitions, such as internal regret. The other one is to change the
reference and compare to the \emph{hypothetical} performance of the best
strategy, in this way accepting a stronger type of 
dependency of the future costs from
the currently selected action (see e.g.\ \cite{Poland:05dule} and the references
therein). It is one of the major open problems to propose refined algorithms and
prove better bounds in this model.

\bibliographystyle{alpha}

\end{document}